\documentclass{article}

\usepackage{PRIMEarxiv}

\usepackage[utf8]{inputenc} 
\usepackage[T1]{fontenc}    
\usepackage{hyperref}       
\usepackage{url}            
\usepackage{booktabs}       
\usepackage{amsfonts}       
\usepackage{nicefrac}       
\usepackage{microtype}      
\usepackage{lipsum}
\usepackage{fancyhdr}       
\usepackage{graphicx}       
\graphicspath{{media/}}     

\title{On the Horizon: Interactive and Compositional Deepfakes
}

\author{
  Eric Horvitz \\
  Microsoft \\
 Redmond, Washington\\
  \texttt{horvitz@microsoft.com}
}

\begin{document}
\maketitle

\begin{abstract}
Over a five-year period, computing methods for generating high-fidelity, fictional depictions of people and events moved from exotic demonstrations by computer science research teams into ongoing use as a tool of disinformation. The methods, referred to with the portmanteau of ``deepfakes," have been used to create compelling audiovisual content. Here, I share challenges ahead with malevolent uses of two classes of deepfakes that we can expect to come into practice with costly implications for society: \emph{interactive} and \emph{compositional} deepfakes. Interactive deepfakes have the capability to impersonate people with realistic interactive behaviors, taking advantage of advances in multimodal interaction. Compositional deepfakes leverage synthetic content in larger disinformation \emph{plans} that integrate sets of deepfakes over time with observed, expected, and engineered world events to create persuasive \emph{synthetic histories}. Synthetic histories can be constructed manually but may one day be guided by  \emph{adversarial generative explanation} (AGE) techniques. In the absence of mitigations, interactive and compositional deepfakes threaten to move us closer to a post-epistemic world, where fact cannot be distinguished from fiction.  I shall describe interactive and compositional deepfakes and reflect about cautions and potential mitigations to defend against them.
\end{abstract}




\section{Introduction}

Democracy depends on an informed and engaged citizenry. Democracy and civil liberties are coming under threat from new forms of disinformation---the distribution of false information with the \emph{intention} of swaying public opinion \cite{horvitz21disinfo}. Fabrications of falsehoods aimed at manipulating masses have a long and tragic history. Highly successful disinformation campaigns have relied on the expertise of propagandists to formulate persuasive false narratives and to propagate verbal and visual information in support of them. Disinformation efforts have grown in sophistication over time, riding on waves of technical advances, from the printing press, to photography, radio and television, and on to internet-based social media, computer graphics, and machine learning. 

We are at an inflection point with the rising capabilities of discriminative and generative AI methods. The advances in machine learning are enabling new forms of content generation and new powers of multimodal interaction. The advances are providing unprecedented tools that can be used by state and non-state actors to create and distribute persuasive disinformation. The rising capabilities frame concerns that our children and grandchildren could find themselves in a post-epistemic world where it is difficult or impossible to distinguish fact from fiction. The speed of these technical developments and new possibilities for their abuse places responsibility in the hands of computer scientists to envision technical futures, likely abuses, and potential mitigations---and to engage across disciplines, organizations, and agencies to raise awareness and collaborate on practices, policies, and regulations. 

Impressive developments in neural models for producing language, visual, and audiovisual content can be exploited for influence operations. Generative neural language models can be directed with ease to synthesize persuasive writings as well as to power compelling dialog aimed at achieving specific goals. Multimodal neural models, which leverage representations constructed from text, images, audio, and video data, can generate realistic visual content based on natural language prompts.  Advances with multimodal neural models include interactive techniques that provide creators with the ability to modify or refine the generated images, using such methods as in-painting, which enables the extension or manipulation of specific regions of images, and prompt engineering, the design of language inputs to produce desired outputs.  While innovations with language-centric and multimodal neural models open up new possibilities for creative expression, imagination, and  education, they can serve as potent weapons of persuasion and disinformation.

AI-generated content about people and events are now being employed in fraud, impersonation, and larger cyber influence programs. Compelling synthetic media, referred to commonly as ``deepfakes," were exotic research projects just a few years ago, first appearing as startling demonstration videos linked with conference papers. Today, open source toolkits are available for producing deepfakes, lowering the bar on required expertise to generate and then distribute them at lightning speed across social media.

We can expect deepfakes to become difficult to discriminate from reality. While numerous methods can be used to generate deepfakes, the challenge with distinguishing fact from fiction is easy to see for the generative adversarial networks (GAN) methodology \cite{goodfellow2014generative}. GANs are an iterative technique where the machine learning and inference employed to generate synthetic content is pitted against systems that attempt to discriminate generated fictions from fact. Both the generator and the detector become increasingly better in the process, with the generator learning over time how best to fool the detector. With this process at the foundation of deepfakes, neither pattern recognition techniques nor humans will be able to reliably recognize deepfakes. 

Turning to the focus of this paper, to date, deepfakes have been constructed and distributed as one-off, stand-alone creations. We can expect to see the rise of new forms of persuasive deepfakes that move beyond fixed, singleton productions \cite{horvitz2022congress}. Malevolent uses of \emph{interactive} and \emph{compositional} deepfakes will leverage advances in the multimodal interaction research community, and wider AI and HCI communities. I will describe these two new expected classes of synthetic media and touch on directions for defending against them. 

\section{Interactive Deepfakes}

\begin{figure*}
\includegraphics[width=\textwidth]{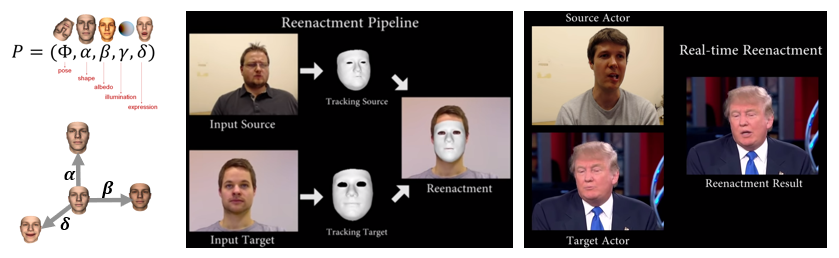}
\caption{Early demonstration of end-to-end pipeline for real-time visual tracking and synthesis of face pose and expression drawn from sets of video frames (derived from \cite{Thies2016}).}
\label{fig:face2face}
\end{figure*}

A constellation of advances in generative AI methods, coupled with frontier research on multimodal interaction, can enable new interactive forms of deepfakes. One of the earliest demonstrations of the use of generative methods for compelling impersonation was presented as an interactive prototype named Face2Face, published at CVPR 2016 \cite{Thies2016}. The project demonstrated how a commodity PC could be used to perform real-time tracking of a \emph{source actor} to control the pose, mouth, and facial expressions of rendered \emph{target actors}, including well-known politicians (see Figure \ref{fig:face2face}). The authors reported that ``the resulting synthesized model is so close to the input that it is hard to distinguish between the synthesized and the real face." More recent generative methodologies can be coupled with real-time tracking to provide similarly rich interactive deepfakes. 

Significant progress on methods for recognition, generation, and interaction pave the way to creating interactive deepfakes. Interactive deepfakes can raise the bar on the persuasiveness of impersonation, bringing new forms of ``presence" and engagement via audio, visual, and audiovisual channels.  Advances in enabling technologies include work on speech recognition, speech production, and appropriate visual renderings of expressions associated with voice activity that enable source actors to have their voice re-rendered as that of the rendered target actors. On voice impersonation, over a decade ago, methods for real-time \emph{voice conversion} were demonstrated, enabling a source speaker to render utterances in another person's voice. More recent advances include the use of deep neural models for generating natural-sounding voices from text \cite{arik2017deep} and the efficient cloning of a target voice from just a few samples of speaking \cite{arik2018neural}. 

An important challenge in multimodal interaction with rendered avatars is generating natural expressions associated with voice activity.  Recent advances in methods for \emph{neural voice puppetry} \cite{thies2020neural} enable compelling real-time generation of appropriate expressions along with synthesized voice. As of 2020, the end-to-end pipeline for mapping audio features of source utterances to a person-specific expression \emph{and} the generation of a photo-realistic rendering took 5ms on an Nvidia 1080Ti \cite{Thies2016}. Beyond human voices, audio capture and re-generation tools, such as WaveNet \cite{oord2016wavenet}, enable the generation of arbitrary background sounds in a deepfake. We are seeing advances in multimodal capabilities for recognition and generation that will soon enable audio, graphical, and language technologies to be woven into toolkits that enable the interactive control of fictional renderings of targeted personalities in teleconferences via the pose, expressions, and voice of a controlling actor.

Scale can be achieved with moving beyond manual control of multimodal impersonation technologies. Various forms of automation can be used to enable a system to convince viewers of the presence of an individual in audio calls and audiovisual conferencing. For example, ambient patterns of motion and attention have long been used in automated visual agents. Expressions, gestures, and other behaviors, such as nods, laughter, applause, and patterns of attention, as captured by the pose and gaze of attendees in a multiparty teleconference call, could be mirrored by an automated avatar based on following and reflecting signals drawn from automated analyses of the identity, speech activity, speech recognition, pose, and expressions of attendees. 

Simple, compelling strategies for projecting the presence of impersonating avatars in audiovisual conference calls include generating the usual greetings and goodbyes of teleconferences. More sophisticated directions with automation can harness rudimentary dialog capabilities that are available today to enable a rendered agent to respond appropriately to specific triggers in the flow of a conversation---such as the call for a vote, agreement or disagreement, or for an opinion which could flow from an agent on command. 

Automated interactive deepfakes could be endowed with basic understandings of the status of flow of a conversation to inform decisions about if and when to interject, leveraging prior work in the multimodal interaction research community on predictive models of when the floor is yielded by a speaker to other participants in a multiparty setting \cite{bohus2011decisions}. The prior work demonstrated the importance of sensing, prediction, and decision-making to guide the fine structure of the timing of turns in multiparty settings. Automated agents must be able to predict the source and target of utterances, and, more generally, the state and dynamics of shifts of the floor in multiparty conversations. Advances in multiparty understanding and interaction will enable new forms of automated impersonation.

Beyond fully manual and automated control of an impersonation, there are opportunities for mixed-initiative approaches, where a source actor can be on standby to take over basic automation of natural patterns of pose and expression as needed when alerted via signals about engagement, complexity, or confusion. With background matching and cloning, a rendering could be slipped into a live conversation without a participant recognizing the substitution, perhaps done briefly enough for an important intervention, such as agreeing with or voting on a proposal. 

Interactive deepfakes can be enhanced in numerous ways with auxiliary audiovisual content, for instance, with introducing simulated or real events occurring in the background that are synchronized with appropriate responses of the impersonating avatar, such as reactions to nearby explosions. The methods could be employed to create persuasive fabricated outcomes, such as the real-time injury of a leader during a call or to convince the public that a leader who has perished is alive and in command.

\section{Compositional Deepfakes}

I use \emph{compositional deepfakes} to refer to a concerning, feasible direction with malevolent uses of synthetic media in influence campaigns on a larger scale: the integration of multiple coordinated deepfakes and/or fabricated events with real-world occurrences to build fictional explanations or \emph{synthetic histories}. Compositional deepfake \emph{plans} include near real-time, as well as the pre- and post-dating one or more deepfakes that project the synthetic history into the past and the future, respectively. Figure \ref{fig:Fig1} depicts a canonical compositional deepfake plan and resulting synthetic history, where a sequence of two fabricated ``past" deepfake media pieces are injected  between two world occurrences and time-stamped as happening at appropriate times between the two events. Moving into the future in this canonical synthetic history, an in-world event is fabricated to complete the persuasive storyline.  As highlighted in the figure, the response to such plans can be monitored and updated with the creation and injection of additional past and future fictional or fabricated events.

\begin{figure*}
\includegraphics[width=\textwidth]{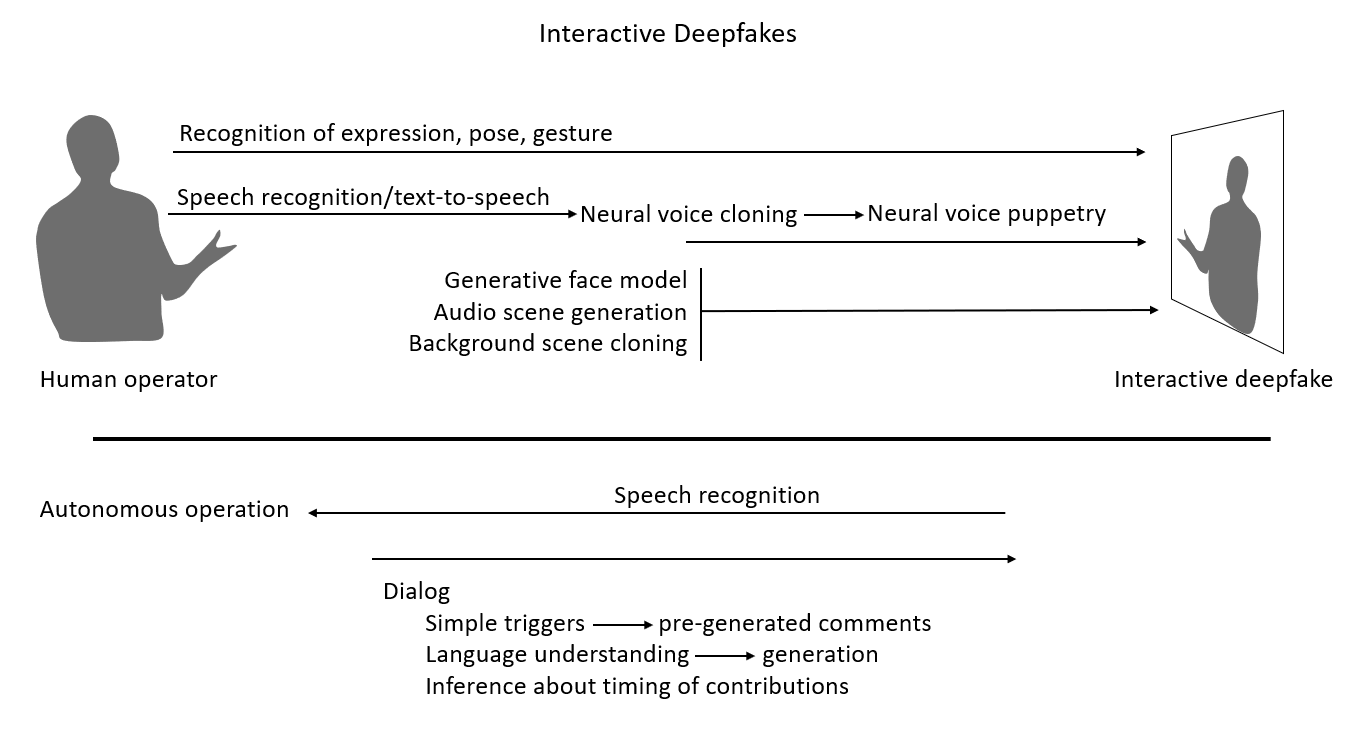}
\caption{Interactive deepfakes. Top: Set of multimodal interaction technologies required for building an interactive deepfake system. Methods include use of high-fidelity renderings with generative AI methods, recognition of pose, expression, and gesture, speech recognition, neural voice cloning, and neural puppetry. Bottom: Opportunities for automation and mixed-initiative control span a range of sophistication, from simple trigger detection and issuance of pre-generated responses to richer models of language understanding and generation.}
\end{figure*}

Compositional deepfakes can be designed to create fictional narratives that are persuasive in their ability to tie together and provide powerful explanations of sets of events in the world to citizens and government leaders. It is not hard to imagine how the explanatory power of custom-tailored synthetic histories could out-compete the explanatory power of the truthful narratives that describe the actual motivations, decisions, and causes of events that are observed in the world. State and non-state actors who seek to generate, execute, and monitor and refine the influences of persuasive compositional deepfakes can harness qualitative models and understandings about how people compose and come to be committed to beliefs about specific explanations. Such understandings include insights and practices developed over many decades to persuade populations, including long-honed nation-state propaganda and more modern commercial sales and marketing campaigns, and recent efforts to optimize engagement and clickthrough with online services that sit at the heart of modern commerce.

\begin{figure*}
\includegraphics[width=\textwidth]{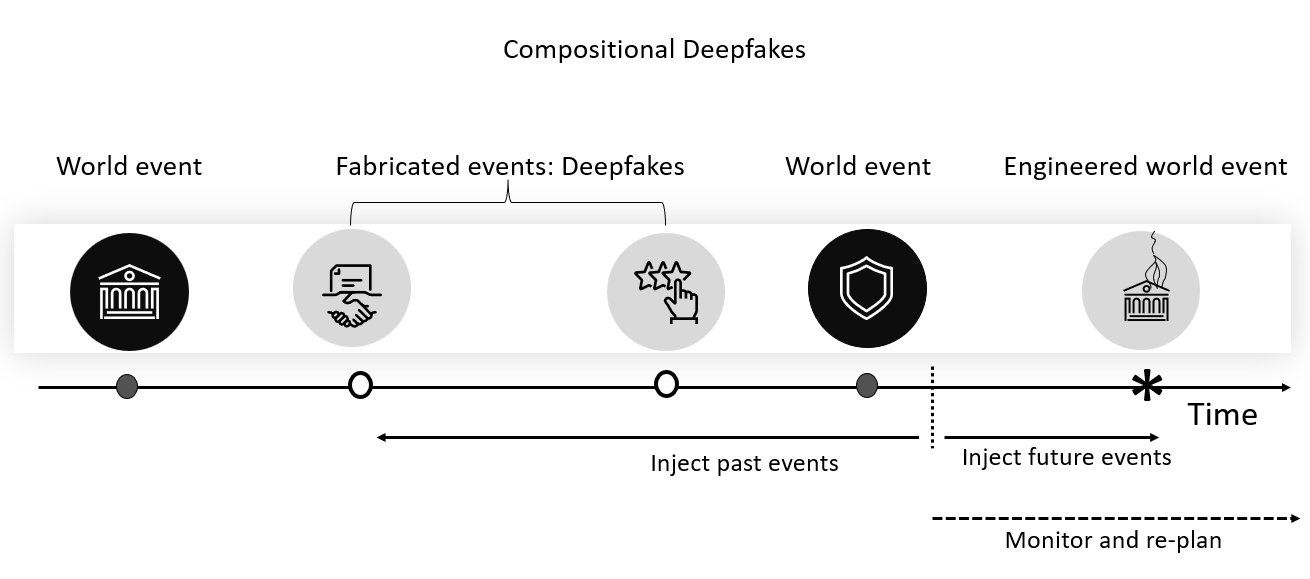}
\caption{Compositional deepfake plans. Persuasive synthetic histories can be created via compositional deepfake strategies that interleave sequences of deepfakes and engineered in-world events with event occurrences in the world. In advance of general releases, testing and refinement can be performed on subpopulations. Follow-on monitoring can allow for fine-grained control of synthetic histories in response to population reactions via modification or extension of synthetic histories. }
\label{fig:Fig1}
\end{figure*}

More fundamentally, compositional deepfake plans can leverage insights about human psychology, including studies of rich sets of biases identified and studied within the cognitive psychology area of judgment and decision making \cite{kahneman1982judgment}. Social and cognitive psychologists have studied how people weave sets of events in the world into convincing narratives. This research includes efforts to understand the formation and commitment of groups to \emph{conspiracy beliefs}, defined as explanations for events or situations attributed to one or more actors working secretively to achieve goals that are unlawful, unfair, or malicious \cite{zonis1994conspiracy}. 

Several investigators have suggested that people are more predisposed to believe in conspiracies when they perceive that a society faces a crisis. They cite times of rapid change, such as the second industrial revolution at the start of the Twentieth Century and the period before World War II, with the internationalization of the economy, new forms of automation, and then the worldwide depression, as examples of this \cite{uscinski2014american}. At such times, populations may feel particularly anxious and insecure, especially citizens who feel that they have little power or voice in decisions \cite{noble1966paranoid}.  These periods of time are ripe for the growth of conspiracy theories that rise to challenge existing political leadership, behavioral norms, and acceptance of groups who may be viewed as outsiders \cite{pipes1999conspiracy}. 

The design palette for compositional deepfakes includes methods for targeting specific individuals or groups, including the creation of multiple narratives, each custom-tailored to different populations. Borrowing from advances in e-commerce, other tools facilitate automated experimentation and refinement on subpopulations in advance of broader releases via testing of the finalized persuasive explanations.  As indicated in the Figure \ref{fig:Fig1}, ongoing monitoring and refinement can be executed via the updating or ``editing" of synthetic histories, where in-world events can be fabricated and deepfakes can be added anywhere on the timeline. 

Other approaches include creating and pre-positioning one or more deepfakes or larger alternate compositional plans into obscure online holding areas at specific dates and following up with deletions and amplifications of content as needed, conditioned on the flow of real-world events. 
Multiple variants of deepfakes can be generated and quietly prepositioned and specific subsets can be brought forward for amplification or deleted depending on events in the world. Forms of pre-positioning have been described as a known disinformation strategy. 

As an example cited as pre-positioning of disinformation \cite{Defending22}, a false claim \cite{Politifact22} of U.S.-funded biolabs in Ukraine being connected to bioweapons development was positioned in a relatively obscure place on YouTube in November 2021 as part of a Russian television series. The story was lifted into prominence at the start of the invasion of Ukraine on February 24th, 2022, when it was simultaneously referred to as a known finding ``from last year" by ten Russian-controlled or highly influenced news sites and then amplified on social media.

Beyond a goal of long-term persuasion, compositional deepfakes can be constructed for use in time-limited operations so as to achieve local goals regardless of whether the fabrications will eventually come to light. For example, the plans can incorporate the temporary disruption of electric power, sensing, or communication aimed at diminishing the ability of people to do real-time fact-checking or discovering evidence of manipulation. Benign explanations for sensing or communication outages can suppress suspicion of a link between an outage and the larger compositional deepfakes.  

\begin{figure*}
\includegraphics[width=\textwidth]{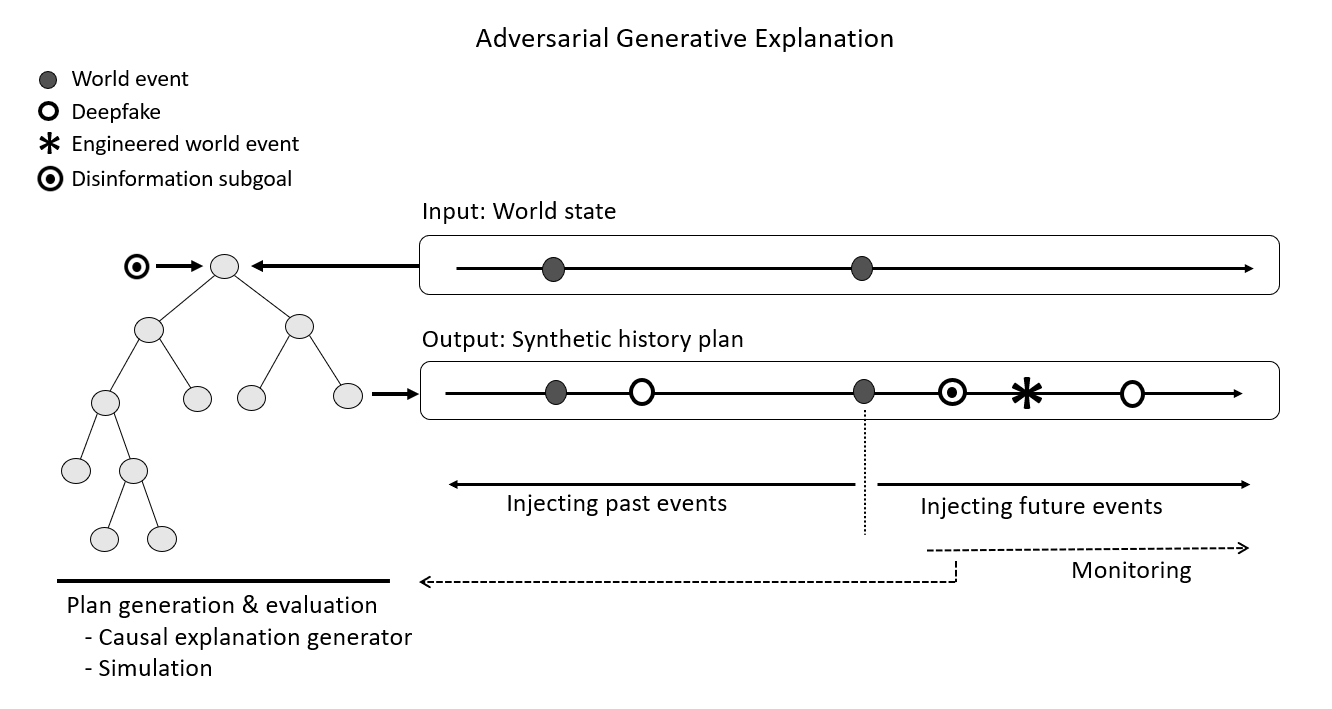}
\caption{Adversarial generative explanation (AGE). Adversarial explanation systems employ causal inference and cognitive models to perform adversarial attacks on understandings of world events. AGE methods depend on search and optimization methods to generate persuasive explanations. The methods take as input key disinformation goals and in-world occurrences to date and generate and test persuasive synthetic histories composed of configurations of synthetic and in-world events. Objective functions for guiding the generation and search can take into consideration the persuasive power of plans and their ability to achieve key disinformation goals. Signals collected via monitoring can guide re-planning with additional content created and injected as prior or follow-on synthetic media and the engineering of additional in-world events.}
\label{fig:Fig2}
\end{figure*}

\section{Adversarial Generative Explanation}

Tragic situations and outcomes for humanity over the course of history are testimony to the ability of expert propagandists to construct and execute persuasive disinformation based on their intuitions and experience. Propagandists have demonstrated how they can weave together real-world observations, fabricated events, and fictional media to create powerful narratives, create and stoke conspiracy beliefs, and achieve goals of moving populations to action---or to acquiescence and inaction \cite{horz2022keep}. Despite the demonstrated capabilities of people to manually author compositional disinformation operations, new forms of assistance and automation is feasible.  

Sketches and more complete synthetic histories, along with recommended sets of disinformation actions, may one day be provided by ``persuasion toolkits." Recent advances in machine learning and inference can power new engines of persuasion in the form of advisory tools or automated services that can generate and persuasive narratives that run counter to truthful explanations. Such systems could be employed for many goals, including political campaigns, legal disputes, and sales and marketing. They could also be aimed at guiding the creation and fielding of sequences of deepfakes and fabrications of real-world events to create persuasive compositional deepfakes as part of generating and target narratives. I refer to these hypothesized tools as \emph{adversarial generative explanation} (AGE) systems. 

AGE systems can be developed by leveraging causal reasoning and psychological models to generate narratives that run counter to truthful explanation of events and intentions. An AGE system could use search and optimization to create persuasive alternative or \emph{contrastive} explanations for sets of observed events in the world. The generation of contrastive, false narratives could take into consideration prior understandings and biases in target populations about the desires, intentions, and actions of people.  Such generations could be guided by specific disinformation goals input as sketches of false narratives or specifications about the actions of specific individuals or organizations as ``must-include" events that are injected into the optimization. 

Beyond building contrastive explanations based on existing or expected real-world events, AGE systems could be harnessed to generate compositional deepfakes, considering ideal sets of synthetic media and fabricated real-world events to introduce as part of generating explanations. Given a set of disinformation goals, recent sets of world events, a production budget, and time horizon, an AGE system could generate alternate synthetic histories, each potentially containing sets of recommended deepfakes and in-world events.  

Figure \ref{fig:Fig2} shows key components of a proposed AGE system, taking into consideration a set of known observations and then countering realistic causal understandings with the generation of adversarial narratives via a process of causal inference, psychological modeling, composition and search. The search can consider different sets of ideal injections of synthetic media and fabricated in-world events to achieve one or more goals, such as promoting a fictional history and narrative persuading populations about the actions and intentions of specific people, organizations, or nation states. 

Causal plans can be generated and evaluated with the assistance of a simulator that uses psychological models in its evaluation. During execution, monitoring can provide input for ongoing revision of the plans, including accretion of new synthetic and fabricated in-world events.  Beyond generating proposals for larger synthetic histories, such tools could provide new forms of ''causal in-painting,” identifying ideal point-wise injections of synthesized events to overlay on real-world events so as to build a plausible story.  

The feasibility of developing powerful AGE systems is supported by recent advances in machine learning, representation, and inference. The systems can draw on methods for performing causal reasoning and explanation \cite{pearl2009causality, spirtes2000causation, halpern2016actual}, including blossoming efforts for integrating causal knowledge into deep learning \cite{causepanel21}. Particularly relevant are recently developed techniques for constructing plausible explanations via multimodal inference that leverages visual, audio, and language capabilities to produce human-understandable causal explanations of events in the world \cite{gerstenberg2021happened} and developments of benchmarks and methods for evaluating causal explanations \cite{roemmele2011COPA}. Such methods could be harnessed to generate and rank adversarial explanations.

Other AI advances that could provide core competencies to AGE systems include deep neural models that are trained to make commonsense ``if-then" inferences about the consequences of actions in the world \cite{sap2019atomic} and about human feelings \cite{rashkin-etal-2018-event2mind}. Relevant advances also come via research on counterfactual reasoning aimed at enhancing explanation of automated inferences. This work spans efforts in machine learning \cite{verma2020counterfactual,van2018contrastive} and Bayesian networks \cite{koopman2021}.  AGE systems might also harness automated inferences about the representation of events in narratives \cite{chambers2008}, including methods for computing expected sequences of events \cite{Sap2022}.

Moving from the computational to psychological realms, the generation and evaluation of candidate adversarial explanations would benefit from research on the psychological front, where researchers have developed qualitative causal models for describing the effectiveness of propaganda based in psychological models \cite{horz2021propaganda, horz2022keep, woodward2005making, lagnado2013causal, lagnado2021explaining, baker2017rational,lombrozo2006functional,kirfel2021inference} and models of persuasion \cite{wood1981stages}.  Other relevant research on the psychological front focuses on gaining understandings of language that can provoke strong emotions \cite{vu2014acquiring}, a factor demonstrated to play a role in the influence of disinformation \cite{martel2020}.

\section{Preparing for Advances in Persuasion and Disinformation}

What might be done to defend against the expected development of integrative and compositional deepfakes? Directions ahead span efforts and innovations in the realms of technology, policy, and practice.

\paragraph{Journalism and reporting.} We need to nurture high-quality journalism and reporting, including the support of local and international news organizations. Efforts include ensuring that trusted reporters are on the ground to observe and record events. The rise of new technologies for impersonation, generating persuasive narratives, and manufacturing synthetic histories will raise the bar on expectations and requirements around reporting. Beyond professional reporting, we need to continue to explore opportunities to engage (and protect) citizen journalists, who can provide multiple, independent signals about world events, including the capture and sharing of photos and audiovisual content from multiple cameras, each with certifiable metadata, such as the location and time of content capture (see the efforts of Witness \cite{Witness22}).

\paragraph{Media literacy.} We will need to foster media literacy and to raise awareness about new forms of manipulation and their growing power to impersonate, fabricate, and persuade. There is evidence that education, including special programs of ``inoculation" and pre-bunking can help can raise alertness and resistance to various disinformation tactics \cite{roozenbeek2022}. Educational programs will have to keep pace with the technical prowess of cyber influence operations, with special attention to raising awareness about the nature and operation of interactive and compositional deepfakes as these methods come into practice. Education and awareness needs to include efforts to disparage truthful explanations that we can expect to come with the rise of powerful disinformation methods: A world of pervasive, persuasive disinformation is conducive to the discrediting of actual happenings--a phenomenon referred to as the ``liar's dividend." Such an approach to discrediting real-world events has been leveraged by malevolent actors who take advantage of common understandings about the ease with which photos can be doctored \cite{Leibowicz21}.

\paragraph{Authenticity protocols.} We will need to stay alert to new forms of generative content and to continue to pursue means for detecting and thwarting inappropriate uses in influence and deception. For example,  special attention will be needed to assuring and asserting the identity of participants in critical private and public meetings.   New authenticity confirming protocols, such as real-time \emph{authenticity challenges}, may need to be introduced to identify interactive deepfakes via required tests of competency and knowledge. New practices of multifactor authentication of identity may become necessary for admittance into online meetings or appearances in videos. 

\paragraph{Content provenance.} We will need to rely increasingly on formal cryptographic pipelines and standards for authenticating the provenance of digital content. Methods and tools for certifying content provenance are recent developments. The methods employ systems and  protocols that certify the source and history of edits made to photos or audiovisual content \cite{england2021amp}.  Digital content provenance methods can raise the level of trust in digital content by helping consumers to understand the organizational source of the content, such as a trusted journalism organization. The methods employ a tamper-proof manifest that travels along with the content which includes the origin of the content and sequence of modifications that may have been made since the publication of the material \cite{england2021amp,aythora2020multi,HorvitzOntheIssues21,C2PA22}. The methods certify that media has not been modified beyond what is indicated in the manifest and that the holder of the cryptographic key created or modified the manifest. Editing and processing without a compliant content provenance pipeline invalidates the manifest. 

Efforts in digital content provenance include work to push authentication closer to reality, with the goal of crytographic ``glass-to-glass" certification---that is, ensuring that the photons hitting the light-sensitive surface of cameras are faithfully rendered as photons on displays. This work includes efforts to build special phones that cryptographically encode time and location along with audiovisual content \cite{Truepic22}. There is work to be done on extending provenance to certifying reality itself. Such work will need to make use of intensive red-teaming aimed at ensuring that the chain of authentication cannot be broken nor gamed.  For interactive deepfakes, it will be important to field real-time versions of digital content authentication. 

Digital content provenance solutions have leveraged private and public distributed digital ledger technologies, including the common blockchain form of ledger. Distributed digital ledgers can provide indelible histories of sequences of media posts. The wide use of public digital ledgers for media will make it difficult to change history with false time stamps for newly posted audiovisual content or other attempts to rewrite the course of events.

\paragraph{Watermarks and fingerprints.} Related to core efforts with digital content provenance are methods that embed in digital content an indelible watermark that withstands well-intentioned and adversarial edits and modifications. Potential watermarks include the use of encoded urls or other coding that point to stored versions of the original content and accessible via privately stored keys. This direction of effort includes the use of soft-hash fingerprints of the content itself that is stored in a database that includes information about the content and its creation.  Work on indelible watermarks and robust fingerprinting could be employed to mark fabricated content as synthetic media. Such watermarking and fingerprinting could be useful for ensuring that synthetic media created for such uses as satire, envisioning, and art are not misinterpreted as capturing real-world events. Indelible watermarking should support signing by creators as well as enable for anonymous publication to protect the authors from retribution by abusive organizations and governments.

\paragraph{Detection.} Research will be needed on the detection and disruption of compositional deepfake campaigns. Detection of compositional plans and synthetic histories will benefit from the development of new tracking tools and ongoing vigilance. Directions include careful monitoring of nation states and organizations with a history of developing disinformation for such behaviors as placing and removing pre-positioned content and sequencing of deepfakes.  Adversarial generative modeling tools may be useful for generating, interpreting, and ``pre-bunking" or reacting to compositional deepfakes.
 
\paragraph{Regulation and self-regulation.} Moving to policy and law, nations and localities will need to consider balanced actions in the regulatory realm aimed at squelching the creation and influence of deepfakes for impersonation and other forms of disinformation, while enabling and protecting free speech. National and international conventions norms, regulations, and laws with stiff penalties may help to stem the tide of new forms of synthetic media aimed at disinformation. However, laws will have to be inspected carefully and be subjected to open debate and ongoing refinement.   In the nearer-term, self-regulation of corporate and academic research labs may be helpful in keeping the most powerful tools and packages out of the hands of malevolent actors. Nonetheless, we must assume that innovations in multimodal interaction, generative AI, and causal modeling and explanation will spread worldwide at lightning speeds and will be harnessed with both beneficent and malevolent intentions. 

\paragraph{Red-teaming and continuous monitoring.} Whether in the realms of technology or policy, threat models and impact assessments will be needed and proposed solutions and mitigations must be carefully ``red-teamed" to ensure robustness of the methods to creative attacks. Well-intentioned innovations aimed at addressing concerns can introduce new avenues for attacks on specific systems and end-to-end operations.  As examples, we need to understand how content reported to come via the phones of multiple citizens might be fabricated and how manifests or watermarks on content might be removed and replaced, and how these and other attacks can lead to confusion or to the bolstering of disinformation.

\section{Conclusion}

We can expect that advances in machine learning and interaction will be leveraged in new forms of persuasion and disinformation. I touched on three concerning directions, including interactive deepfakes, compositional deepfakes, and adversarial generative explanation. Computer scientists innovating at the frontiers are uniquely qualified to anticipate scientific innovations and to kick off envisioning across multiple disciplines about how the advances may be harnessed, both for the greater good and for malevolent ends. Vigilance will be needed on potential uses of our models, data, results, and technologies for disinformation. As we progress at the frontier of technological possibilities, we must continue to envision potential abuses of the technologies that we create and work to develop threat models, controls, and safeguards---and to engage across multiple sectors on rising concerns, acceptable uses, best practices, mitigations, and regulations.

\section*{Acknowledgments}
I thank Neil Coles, Paul England, Andrew Jenks, Ram Shankar Siva Kumar, Sarah McGee, Georgianna Shea, Allison Stanger, Subramaniam Vincent, Rand Waltzman, and Ben Zorn for their feedback on an earlier version of the manuscript.  



\bibliographystyle{ACM-Reference-Format}
\bibliography{sample-base}










\end{document}